%% file: ieee_main.tex
\documentclass[letterpaper, 10 pt, conference]{ieeeconf}  
\usepackage[T1]{fontenc}
\usepackage{algorithm2e}
\RestyleAlgo{ruled}
\usepackage{xcolor}
\usepackage{upgreek}
\usepackage{multicol}
\usepackage{dsfont,setspace}
\usepackage{amsmath,graphicx}
\usepackage{float}
\usepackage{amssymb}
\usepackage{amsmath}
\usepackage[normalem]{ulem}
\usepackage[thinc]{esdiff}

\usepackage{lineno}
\usepackage{epstopdf}
\usepackage{authblk}
\usepackage{dblfloatfix}
\usepackage{tabularx}
\usepackage{soul}

\usepackage[
  separate-uncertainty = true,
  multi-part-units = repeat
]{siunitx}

\usepackage{subcaption}
\usepackage[colorlinks]{hyperref}
\hypersetup{
     colorlinks,
     linkcolor=black,
     citecolor=black,
     filecolor=black,
     urlcolor=blue,
 }
\usepackage{cleveref}
\usepackage{textcomp}
\usepackage{gensymb}
\usepackage{tikz}
\usepackage{pifont}
\usepackage{makecell}
\usepackage{svg}

\newcommand{\cmark}{\ding{51}}%
\newcommand{\xmark}{\ding{55}}%

\SetKwComment{Comment}{/* }{ */}
\SetKwInput{KwRequire}{Require}

\newcommand{\grey}[1]{
{\color{gray}{#1}}
}

\newcommand{\red}[1]{
{\color{red}{#1}}
}

\newcommand{\green}[1]{
{\color{teal}{#1}}
}

\IEEEoverridecommandlockouts                              

\begin{document}

\title{\LARGE \bf PIP-Loco: A Proprioceptive Infinite Horizon Planning Framework for Quadrupedal Robot Locomotion}

\author{
Aditya Shirwatkar$^{*1}$, Naman Saxena$^{*1}$, Kishore Chandra$^{1}$, Shishir Kolathaya$^{2}$
\thanks{$^*$These authors have contributed equally. This work is funded by Kotak IISc AI-ML Centre (KIAC) and Army Design Bureau.}
\thanks{$^{1}$A. Shirwatkar, N. Saxena and K. Chandra are with the Robert Bosch Center for Cyber-Physical Systems, Indian Institute of Science, Bengaluru.}%
\thanks{$^{2}$S. Kolathaya is with the Robert Bosch Center for Cyber-Physical Systems and the Department of Computer Science \& Automation, Indian Institute of Science, Bengaluru.}
\thanks{Project Website: \href{https://www.stochlab.com/PIP-Loco/}{stochlab.com/PIP-Loco/}, Email: \href{mailto:stochlab@iisc.ac.in}{stochlab@iisc.ac.in}}%
}

\maketitle
\thispagestyle{empty}
\pagestyle{empty}

\begin{abstract}
A core strength of Model Predictive Control (MPC) for quadrupedal locomotion has been its ability to enforce constraints and provide interpretability of the sequence of commands over the horizon. However, despite being able to plan, MPC struggles to scale with task complexity, often failing to achieve robust behavior on rapidly changing surfaces. On the other hand, model-free Reinforcement Learning (RL) methods have outperformed MPC on multiple terrains, showing emergent motions but inherently lack any ability to handle constraints or perform planning. To address these limitations, we propose a framework that integrates proprioceptive planning with RL, allowing for agile and safe locomotion behaviors through the horizon. Inspired by MPC, we incorporate an internal model that includes a velocity estimator and a Dreamer module. During training, the framework learns an expert policy and an internal model that are co-dependent, facilitating exploration for improved locomotion behaviors. During deployment, the Dreamer module solves an infinite-horizon MPC problem, adapting actions and velocity commands to respect the constraints.
We validate the robustness of our training framework through ablation studies on internal model components and demonstrate improved robustness to training noise. Finally, we evaluate our approach across multi-terrain scenarios in both simulation and hardware. 
\end{abstract}


\input{sections/introduction}

\input{sections/methods}

\input{sections/training}
\input{sections/results}

\input{sections/conclusion}

\bibliographystyle{ieeetr}
\footnotesize
\bibliography{references}

\end{document}

%% file: sections/introduction.tex
\section{Introduction}

\begin{figure}[htp!]
    \captionsetup{font=small}
    \centering
    \includegraphics[width=0.6\linewidth]{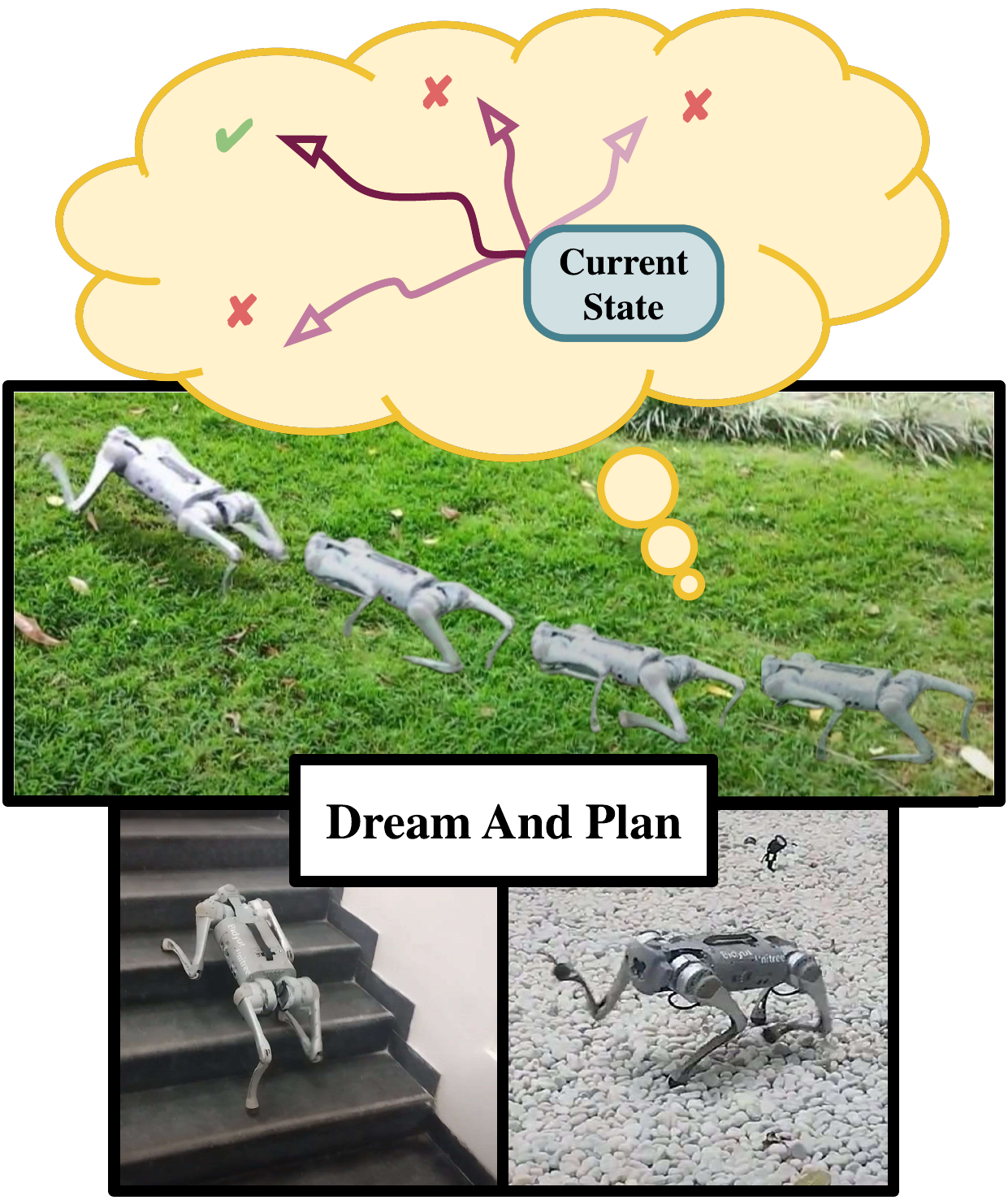}
    \caption{\textit{Conceptual overview of PIP-Loco:} The framework plans actions and velocity commands by dreaming about future states and enforcing constraints. The top shows the potential future trajectories starting from the current state, while the bottom depicts the quadruped adapting to stairs and rocky terrain.}
    \vspace{-18pt}
    \label{fig:overview}
\end{figure}

Model Predictive Control (MPC) methods \cite{convexmpc,perceptmpc,mpc2,mpc3,jumpmpc} have been the default choice for designing controllers for quadruped robots for a very long time. This is because of their ability to enforce various domain/task-specific constraints to produce safe and agile behaviors. Moreover, they offer interpretability of optimized control commands by being able to plan. \cite{convexmpc} simplified the task of obtaining a quadruped controller by formulating it as a convex optimization problem. \cite{jumpmpc} used nonlinear MPC to achieve complex motions such as jumping, bounding, and trotting. Despite their success, MPC methods are less versatile in cluttered environments as they make strict assumptions about the model (like single rigid body dynamics or inverted pendulums). In addition, the end-foot trajectories are generated from a limited class, where the focus is typically on identifying the take-off and landing points. All these issues act as a deterrent for realizing truly emergent behaviors.

\begin{table*}
    \small
    \captionsetup{font=small}
    \centering
    \caption{\textit{Framework Comparisons:} The adaptation module models environment hidden parameters (like friction, contact forces, etc). An internal model predicts the robot base velocities, latent states for terrain characteristics/disturbance responses, or future observations up to $H$ steps. NLM, PLM, and FLM denote the No-Latent, Partially-Latent, and Fully-Latent Modules, respectively (see Section \ref{sc:internal_model}).}
        \label{tab:comparison_table}
    \setlength{\arrayrulewidth}{1.5pt}
    \begin{tabular}{c c c c c c}
        \hline
        \thead{\textbf{Method}} & \thead{\textbf{Observation} \\ \textbf{History Size}} & \thead{\textbf{Adaptation} \\ \textbf{Module}} & \textbf{Internal Model} & \textbf{Multi-terrain} & \thead{\textbf{Planning} \\ \textbf{(Deployment)}} \\
        \hline \\[-1em]
        Walk-these-ways \cite{wtw} & 30 & latent & $-$ & \red{\xmark} & \red{\xmark} \\
        RMA \cite{rma} & 50 & latent & $-$ & \green{\cmark} & \red{\xmark} \\
        DreamWaQ \cite{dreamwaq}, HIMLoco \cite{himloco} & 6 & $-$ & latent ($H=1$) & \green{\cmark} & \red{\xmark} \\
        PIP-Loco w/ NLM (Ours) & 1 & $-$ & observations ($H\geq 1$) & \green{\cmark} & \green{\cmark} \\
        PIP-Loco w/ PLM (Ours) & 6 & $-$ & \makecell{observations + latent \\ ($H\geq 1$)} & \green{\cmark} & \green{\cmark} \\
        PIP-Loco w/ FLM (Ours) & $H$ & $-$ & latent ($H\geq 1$) & \green{\cmark} & \green{\cmark} \\
        \hline
    \end{tabular}
    \vspace{-10pt}
\end{table*}

On the other hand, model-free reinforcement learning (RL) methods have produced more robust and adaptive controllers compared to MPC methods. Such RL-based controllers have demonstrated significant potential for enabling quadruped locomotion across diverse terrains, including flat surfaces, stairs, slopes, and slippery environments \cite{eth2019, eth2020, eth2022}. The introduction of massively parallel RL agent training in IsaacGym \cite{leggedgym} has further accelerated the development of RL controllers for quadrupeds. By leveraging this parallelism, \cite{wtw} successfully designed a controller capable of adjusting various gait parameters, such as body height, step frequency, and step height, using Raibert's heuristic. Additionally, the works of \cite{parkourcmu, parkoureth, parkourstanford} have showcased RL's ability to enable quadrupeds to perform parkour-like maneuvers. 

Despite these advancements, several challenges still need to be solved when designing RL controllers. A critical issue is the reliable observability of essential states, such as the robot's linear velocity, which affects both model-free and model-based approaches that typically combine state estimators to mitigate sensor noise.
For this,  \cite{himloco} proposed HIM-Loco that addresses noisy hardware observations by employing an internal model, drawing inspiration from \cite{dreamwaq} 's use of the asymmetric actor-critic framework. HIM-Loco demonstrated robustness across various terrains, such as sandy slopes, stairs, and bushes. However, their approach only considered one-step latent state prediction within the internal model and did not explore extending this model further. Training across diverse terrains simultaneously also often leads to policies that give generalized yet suboptimal actions. During deployment on a specific terrain, these actions can be improved through planning to achieve a more appropriate local optimum. 
Thus, taking inspiration from MPC and Dreamer World Models \cite{dreamer, daydreamer}, in this paper, we aim to advance the internal model techniques and address challenges like constraint satisfaction and interpretability of behaviors by integrating planning mechanisms. 

\begin{figure*}[htp!]
    \captionsetup{font=small}
    \centering
    \begin{subfigure}{0.5\textwidth}
        \centering
        \includegraphics[width=\linewidth]{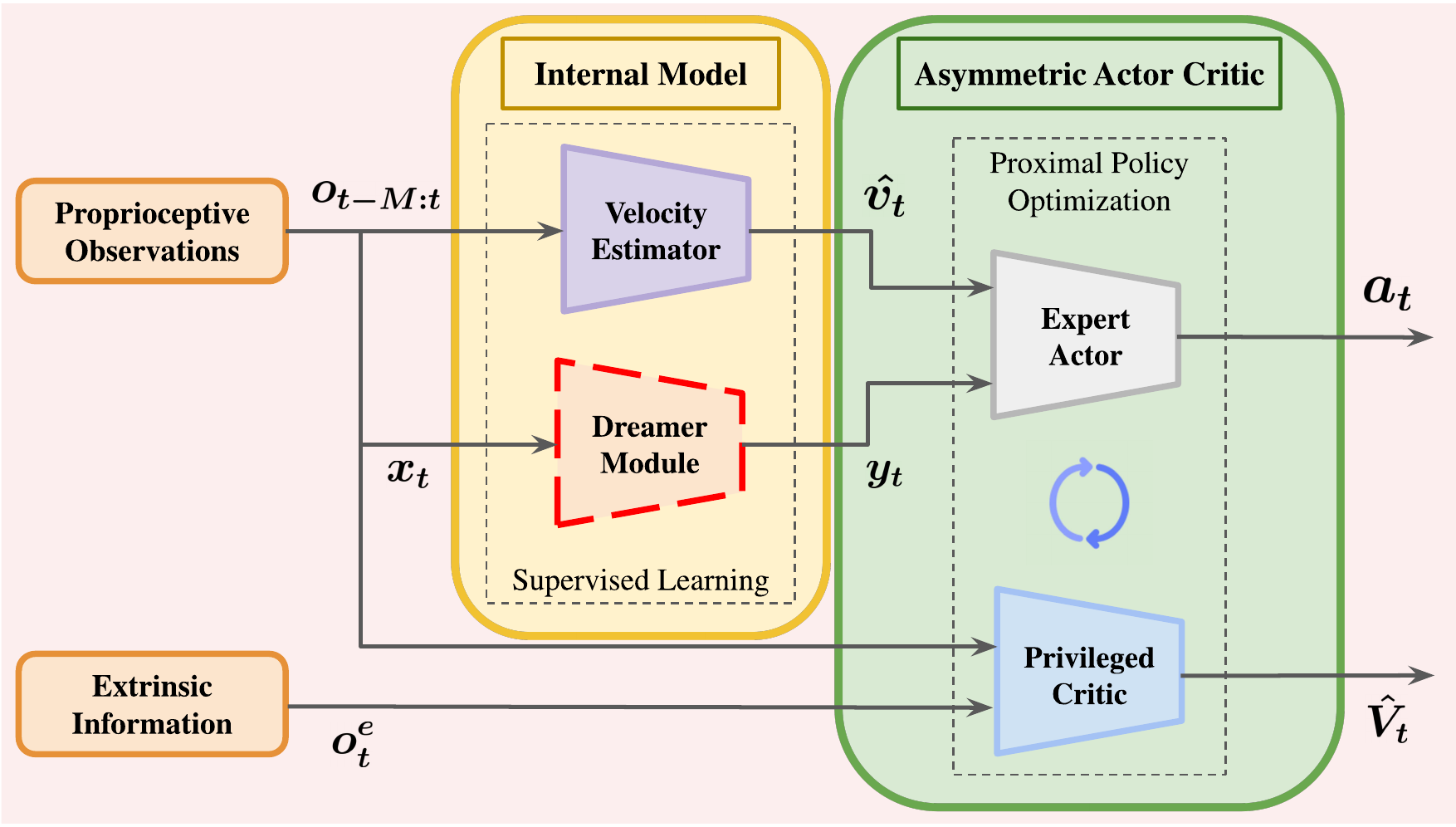}
        \caption{Training}
        \label{fig:training}        
    \end{subfigure} %
    \begin{subfigure}{0.02\linewidth}
        \centering
        \hspace{1pt}
        \tikz{\draw[-,gray, dashed, ultra thick](0,0) -- (0,4.8);}
        \vspace{18pt}
    \end{subfigure} %
    \begin{subfigure}{0.21\textwidth}
        \centering
        \includegraphics[width=\linewidth]{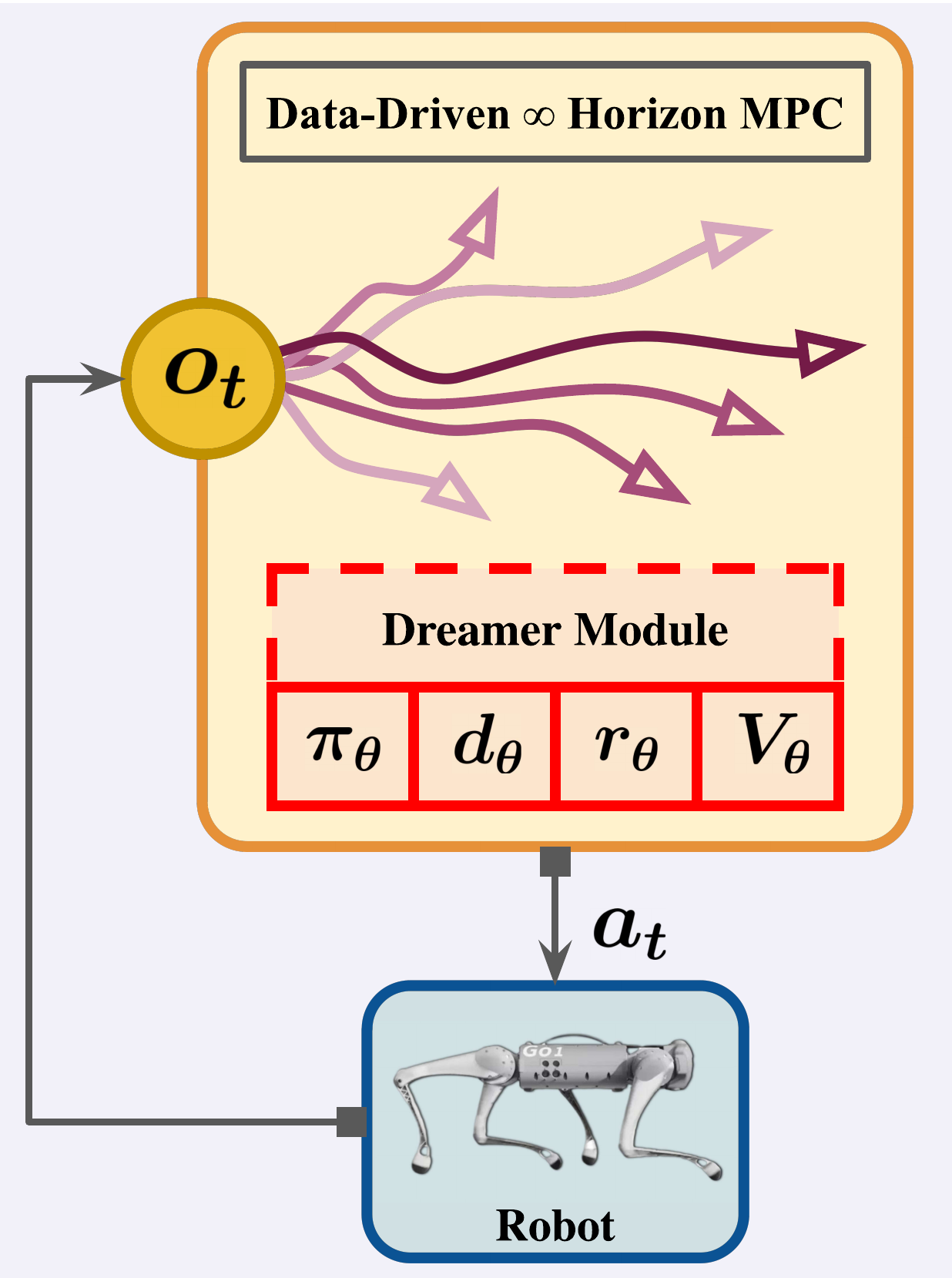}
        \caption{Deployment}
        \label{fig:deployment}
    \end{subfigure}
    \caption{\textit{PIP-Loco Framework}: (a) The internal model (comprising of a velocity estimator and Dreamer module) learns in a co-dependent way with the Asymmetric Actor-Critic. The Dreamer module facilitates temporal reasoning by dreaming about future observations and latent states, enhancing exploration for improved locomotion behaviors. (b) The Dreamer module solves an infinite-horizon MPC problem to generate actions \( a_t \) for the robot, ensuring robust constraint handling and adaptive locomotion across terrains.} 
    \label{fig:diagram}
    \vspace{-15pt}
\end{figure*}

We propose a proprioceptive infinite-horizon planning framework for quadruped locomotion, termed \textit{PIP-Loco} (see Fig. \ref{fig:overview}). While training, we learn an expert policy (actor) and an internal model (comprising a velocity estimator and a Dreamer module) in a co-dependent manner. The Dreamer module includes a policy and value function cloned from the expert actor and privileged value function (critic), respectively, a reward function and a dynamics model to provide future states. In deployment, the Dreamer module solves an infinite-horizon optimization problem in an MPC fashion. This planning mechanism facilitates the filtering and adaptation of actions and velocity commands that might otherwise violate constraints, leading to robust quadruped locomotion. 
Table \ref{tab:comparison_table} compares our work with the state-of-the-art. While we analyze multiple variants of the Dreamer module to assess the impact of increasingly complex internal models, our deployment relies on the simplest No-Latent Module (NLM) variant (see \ref{sc:internal_model}). The following are the broad contributions of our work:
\begin{enumerate}
    \item We introduce a proprioceptive infinite horizon planning framework for RL in quadrupedal locomotion for the first time. This is enabled through learning an internal model that incorporates dreaming about future observations.
    \item Further, we study three types of internal models to understand the importance of latent and future observations.
    \item We deployed our proposed method on the Unitree Go1 robot to demonstrate constraint satisfaction and robustness of locomotive behaviors on multiple terrains.
\end{enumerate}

The benefits of our proposed framework are threefold. The first is overcoming limited observability in real-world settings through an internal model. The second is providing interpretability by leveraging the Dreamer module to simulate future trajectories, allowing for inspection and correction of predicted robot behaviors. Lastly, enabling the incorporation of hard constraints, such as joint limits or body stability conditions, directly into the planning process during deployment, ensuring safety and feasibility.

A detailed description of our approach is provided in Section \ref{sc:method}, with quantitative results from the Unitree Go1 robot and ablation studies of different internal model configurations and planning presented in Section \ref{sc:results}.

%% file: sections/methods.tex
\section{methodology} \label{sc:method}
This section will describe our proposed method, starting with some preliminary concepts. Later in Section \ref{sc:why_not_offpol}, we will discuss the motivation for using the expert actor for interaction during training and the Dreamer module in deployment.

\subsection{Preliminaries}
We model quadruped locomotion as an infinite-horizon discounted Partially Observable Markov Decision Process (POMDP), defined by the tuple $\mathcal{M} = \{\mathcal{S}, \mathcal{A}, \mathcal{O}, r, \mathcal{P}, \gamma, \mathcal{F}\}$. Here, $\mathcal{S} \subset \mathbb{R}^n$ is the state space, $\mathcal{O} \subset \mathbb{R}^p$ is the observation space, $\mathcal{A} \subset \mathbb{R}^m$ is the action space, and $\mathcal{O}^M \subset \mathbb{R}^{p \times M}$ denotes the observation history up to $M$ steps. The mapping $\mathcal{F}: \mathcal{S} \mapsto \mathcal{O}$ provides observations. The reward function is $r : \mathcal{S} \times \mathcal{A} \mapsto \mathbb{R}$, and the transition kernel is $P: \mathcal{S} \times \mathcal{A} \mapsto \Pr(\cdot)$, with $\Pr$ a probability measure. Finally, the discount factor is $\gamma \in (0, 1)$. We solve this POMDP task using the concept of Asymmetric Actor-Critic \cite{asymmetric}, where the input to the expert actor differs from the input to the privileged critic. 

\subsection{Training}
While the expert actor and privileged critic are trained via the Proximal Policy Optimization (PPO) \cite{ppo}, the internal model is trained via supervised learning (see Fig. \ref{fig:training}). The choice of input to the expert actor depends on the internal model used. However, the observation space (excluding the internal model output), action space, and reward function remain the same across the different models studied.

\subsubsection{Observation Space}
At each time step $t$, the observation $o_t$ consists of the robot's joint angles relative to their nominal values ($q_t - q_{\text{nominal}}$), joint velocities ($\dot{q}_t$), projected gravity ($g$), target velocity commands for the robot's base ($\text{v}^{cmd}_{xy}, \omega_z^{cmd}$), and the previous action ($a_{t-1}$).

\subsubsection{Action Space}
The policies produce joint-angle perturbations ($a_t$) relative to the nominal joint angles ($q_{\text{nominal}}$) for all 12 joints of the quadruped. The joint angle command, defined as $q_t = a_t + q_{\text{nominal}}$, is sent to an actuator network \cite{eth2019} to produce the desired torques. 

\subsubsection{Reward Function}
The reward function integrates several terms to encourage desired behaviors, such as achieving target linear and angular velocity commands, maintaining body height, and preserving orientation. We refer the reader to \cite{himloco} for a detailed description of the specific reward terms. Additionally, we introduce a barrier function-based reward term to promote smooth motion and reduce abrupt or aggressive movements, as discussed in \cite{barrier}.

\subsubsection{Internal Model}\label{sc:internal_model}
The concept of an internal model, introduced in \cite{dreamwaq,himloco}, leverages an encoder network in conjunction with a velocity estimator ($v_\phi$). This output is provided to an RL policy (expert actor) within an Asymmetric Actor-Critic framework. The internal model improves the expert actor's resilience to sensor noise and allows it to infer terrain characteristics or simulate disturbance response in latent space. However, the optimal design of such internal models and the importance of dreaming about future observations or its application in planning remain unexplored.

Hence, in this work, we introduce an internal model that incorporates a velocity estimator, denoted as $v_\phi: \mathcal{O}^M \mapsto \mathbb{R}^3$, alongside three distinct variations of the Dreamer module. Each of these variants includes a dynamics model ($d_\theta$), a cloned policy model ($\pi_\theta$), a reward model ($r_\theta$), a value function ($V_\theta$), and an optional encoder ($\alpha_\theta$). While these variants differ in architectural design, they share the same core components. Our motivation for studying these variants stems from the need to identify an effective "dreaming" process within the Dreamer module that can be successfully applied to solve an MPC problem during deployment. Below, we provide a detailed description of each variant, followed by an evaluation of their performance in Section \ref{sc:which_internal_model}.

\paragraph{No-Latent Module (NLM)}
The No-Latent Module (NLM) operates without relying on any latent representations. Instead, it directly uses current observation ($x_t = o_t$), to predict future observations over a horizon $H$ using a dynamics model $d_\theta^{NLM} : \mathcal{O} \times \mathcal{A} \mapsto \mathcal{O}$, as shown in \eqref{eq:internal_nlm}. Here, we use $\Tilde{o}$ to denote the observations predicted during dreaming. The behavior-cloned policy, $\pi_\theta^{NLM}: \mathcal{O} \mapsto \mathcal{A}$, generates actions, while the reward model  $r_\theta^{NLM}: \mathcal{O} \times \mathcal{A} \mapsto \mathbb{R}$ and value function $V_\theta^{NLM} : \mathcal{O} \mapsto \mathbb{R}$ predict expected rewards and future returns. Due to the simplicity of its inference process and its competitive performance compared to other variants (see results in Section \ref{sc:which_internal_model}), we utilized NLM for planning during the deployment stage.

{\small \vspace{-5pt} \begin{equation}\label{eq:internal_nlm}
\begin{aligned}
    y_{t}^{NLM} = \big \{\Tilde{o}_{t:t+H} \ & \big| \ \Tilde{o}_{k+1} = d_\theta^{NLM}(\Tilde{o}_k, \pi_\theta^{NLM}(\Tilde{o}_k)) \\ & \Tilde{o}_t = o_t, \ \forall \ k = t, \cdots, t+H-1 \big \}.
\end{aligned}
\end{equation}}

\paragraph{Partially-Latent Module (PLM)}
The Partially-Latent Module (PLM) extends the NLM by using a history of observations ($x_t = o_{t-M:t}$) to capture temporal dependencies. The output ($y_t^{PLM}$) consists of predicted future observations over a horizon of length $H$, as outlined in \eqref{eq:internal2}. It also incorporates an encoder $\alpha_\theta^{PLM}: \mathcal{O}^{M} \mapsto \mathbb{R}^q$, which compresses the observation history into a latent vector ($z_t$) to improve predictions. The dynamics model $d_\theta^{PLM}: \mathcal{O} \times \mathbb{R}^q \times \mathbb{R}^3 \times \mathcal{A} \mapsto \mathcal{O}$ uses the latent vector, current observation, velocity estimates, and actions to predict the future observation. The reward model $r_\theta^{PLM}: \mathcal{O} \times \mathbb{R}^q \times \mathbb{R}^3 \times \mathcal{A} \mapsto \mathbb{R}$, and value function $V_\theta^{PLM}: \mathcal{O} \times \mathbb{R}^q \times \mathbb{R}^3 \mapsto \mathbb{R}$ similarly also leverage the latent representation and velocity estimates to predict expected rewards and future returns. Actions are generated by the policy $\pi_\theta^{PLM}: \mathcal{O} \times \mathbb{R}^q \times \mathbb{R}^3 \mapsto \mathcal{A}$, using observation, latent vectors, and velocity estimates.

{\small \vspace{-5pt} \begin{equation}\label{eq:internal2}
\begin{aligned}
    y_{t}^{PLM} = \big \{(\Tilde{o}_{t:t+H}&, \ z_t) \\ \big | \ \Tilde{o}_{k+1} = d_\theta^{PLM}&(\Tilde{o}_k, z_k, v_k, \pi_\theta^{PLM}(\Tilde{o}_k, z_k, v_k)), \\ z_k =  \alpha&_\theta^{PLM}(\Tilde{o}_{k-M:k}), v_k = v_\phi(\Tilde{o}_{k-M:k}) \\ & \Tilde{o_t} = o_t, \ \forall \ k = t, \cdots, t+H-1 \big \}.
\end{aligned}
\end{equation}}

\paragraph{Fully-Latent Module (FLM)}
The FLM is similar to PLM, except for an extra step of passing the output of PLM through the encoder $\alpha_\theta^{PLM}$ to compress the future information of horizon length $H$ into a latent vector ($y_t^{FLM} = \{\alpha_\theta^{PLM}(\Tilde{o}_{t:t+H}),\alpha_\theta^{PLM}(o_{t-H:t})\}$). Note that in FLM, we set $M=H$, a design choice that simplifies the architecture by enabling the use of a single encoder. Lastly, the richer information in PLM and FLM variants enhances decision-making in environments with strong temporal dependencies. However, this comes at the cost of increased computational overhead compared to the NLM.

The velocity estimator network ($v_\phi$) is trained using mean squared error (MSE), with the robot's actual velocity as the ground truth. Similarly, both the dynamics model ($d_\theta$) and reward model ($r_\theta$) are optimized using MSE. The encoder ($\alpha_\theta$) employs contrastive learning \cite{himloco} to ensure informative latent representations. The policy network ($\pi_\theta$) and value function ($V_\theta$) within the Dreamer module are trained via behavior cloning and supervised learning: $\pi_\theta$ imitates an expert actor ($\pi_e$), while $V_\theta$ is trained to match a privileged critic ($V_e$).

\subsubsection{Asymmetric Actor Critic}\label{sc:aac}
Many previous works (\cite{rma,eth2020}) have used adaptation modules to explicitly learn to encode privileged observations through regression. However, these adaptation modules typically require two separate training phases and are limited by approximation errors. In contrast, with our framework, the asymmetric actor-critic implicitly encourages the internal model outputs fed to the expert actor to encode information from privileged observations and enable dreaming. 

The expert actor ($\pi_e$) receives the following inputs: the current observation ($o_t$), the current velocity estimate ($\hat{v}_t$), and the output of the Dreamer module ($y_t$). In our work, which builds on the HIM-Loco framework \cite{himloco}, the expert actor is trained to integrate these inputs effectively. On the other hand, the privileged critic ($V_e$) takes as input both the current observation ($o_t$) and the privileged observation ($o^{e}_t$) obtained from the simulator. This privileged observation includes information such as the height field of the surface around the robot, external forces acting on the robot, and the robot's linear velocity, among other data.

As shown in Fig. \ref{fig:training}, a key distinction of our training approach is that the Dreamer module and the expert actor are interdependent, as both are learned simultaneously in a single phase. This co-dependence exists because the quality of $y_t$ relies on the actions predicted by $\pi_\theta$, which, in turn, depends on the performance of $\pi_e$.

In this section, we have detailed the training process conducted within the simulator. Moving forward, we will delve into how to use these Dreamer modules (specifically NLM) during deployment. The ensuing discussion will highlight the practical considerations and modifications required for effective deployment in real-world environments.

\subsection{Deployment}\label{sc:planning}
As discussed earlier, an online planner is essential for filtering out unsafe or suboptimal actions generated by the RL policy. It also enhances the interpretability of the robot's behavior by providing insights into the agent's dreaming process. To achieve this, we incorporate infinite-horizon planning inspired by techniques such as \cite{loop, tdmpc, tdmpc2}. Our approach, as shown in Fig. \ref{fig:deployment}, involves solving the following optimization problem to obtain the optimal action sequence $a^*_{t:t+H}$ and a safe twist command $\nu^*_{cmd}:= (\text{v}^{cmd}_{xy}, \omega^{cmd}_z)$ at the current observation $o_t$,

{\small \vspace{-10pt} \begin{equation}\label{eq:inf_mpc}
\begin{aligned}
    \underset{\nu_{cmd}, \ (a_t, \cdots, a_{t+H})}{\max } \ & \sum_{k=0}^{H-1} \gamma^k r_\theta(\hat{o}_{t+k}, a_{t+k}) + \gamma^H V_\theta(\hat{o}_{t+H}),\\
    \text{s.t. } & c_k(\hat{o}_{t+k}, a_{t+k}, \nu_{cmd}) \leq b_k, \\
    \hat{o}_{t+k+1} = & \ d_\theta(\hat{o}_{t+k}, a_{t+k}), \ \forall \ k = 0, \cdots, H-1.
\end{aligned}
\end{equation}}

In this formulation, $\hat{o}_t$ denotes the observation where components corresponding to the velocity command in $o_t$ are replaced with the optimization variable $\nu_{cmd}$. $\nu_{cmd}$ may vary for each instance of $k$; however, for the sake of simplicity, it is assumed to remain consistent throughout the planning horizon.
The parameter $\gamma$ represents the discount factor, and $c_k(\cdot) \leq b_k$ specifies the constraints that must be satisfied at each step $k$.

{
\begin{algorithm}[htp!]
\small
\caption{MPC Planner (\textit{deployment})}\label{alg:mpc}
\KwRequire{
$\theta$: Dreamer module parameters (NLM variant); \ \ \ \ \ \
$(\mu_{t-1}, \sigma_{t-1})$: previous iteration's $\mathcal{N}$ parameters;
$o_t$: Current observation; \ \ \ \ \ \ \ \ \ \ \ \ \ \ \ \ \ \ \ \ \ \ \ \ \ \
$\nu^{tgt}_{cmd}$: Target twist command;
}

\If{t == 0}{
    Set the warm start $(\mu_{t-1}, \sigma_{t-1})$ with $\pi_\theta$ using $d_\theta$.
}

Set $(\mu^0, \sigma^0) \leftarrow (\mu_{t-1}, \sigma_{t-1})$\;

\For{$i=1$ \textbf{to} \textit{N}}{
    Init. empty sets $A^E_\tau, A^S_\tau$ and $R_\tau$\;

    Get $\hat{o_t}$ from $o_t$ and $\nu^i_{cmd}$\;
    
    Collect $M_\pi$ command and action traj. of length $H$ using $\pi_{\theta}$ and dynamics model $d_{\theta}$\;
    
    Collect $M$ command and action traj. using $\mathcal{N}(\mu^{i-1}, \sigma^{i-1})$\;

    \For{each $(\nu^{j}_{cmd}, a^j_t, a^j_{t+1}, \cdots, a^j_{t+H})$\ \ \ \ \ \ \ \ \ \ \ \ \ \ \ \ \ \ from j = 0 \textbf{to} $M + M_\pi$}{
        
        $R_j, C_j = 0, 0$\;
        
        \For{k = 0 \textbf{to} $H-1$}{
            $R_j = R_j + \gamma^k r_\theta(\hat{o}^j_{t+k}, a^j_{t+k})$\;

            $C_j = C_j + \gamma^k c(\hat{o}^j_{t+k}, a^j_{t+k}, \nu^{j}_{cmd})$\;
            
            $\hat{o}^j_{t+k+1} = d_{\theta}(\hat{o}^j_{t+k}, a^j_{t+k})$\;
        }
        
        $R_j = R_j + \gamma^H V_\theta(o^j_{H}) - \lambda C_j$\;

        \If{$C_j \leq b$}{
            $A^S_\tau \leftarrow A^S_\tau \cup (\nu^{j}_{cmd}, a^j_t, a^j_{t+1}, \cdots, a^j_{t+H})$\;   
            $R_\tau \leftarrow R_\tau \cup R_j$\;
        }
    }

    Sort $A^S_\tau$ according to their returns $R_\tau$\;

    $A^E_\tau \leftarrow$ select top-$M_{elite}$ traj. from $A^S_\tau$\;

    \CommentSty{\grey{// MPPI Update}}
    
    $(\mu_{elite}, \sigma_{elite}) \leftarrow$ fit $\mathcal{N}$ to the elite set $A^E_\tau$\;

    $(\mu^i, \sigma^i) = \beta (\mu_{elite}, \sigma_{elite}) + (1-\beta) (\mu^{i-1}, \sigma^{i-1})$\;
    
}
\KwResult{$\nu_{cmd}$ \& first action sampled from $\hspace{-1.8pt}\mathcal{N}(\mu^N, \sigma^N)$}
\end{algorithm}

\begin{table}[htp!]
    \scriptsize
    \centering
    \caption{Tunable hyperparameters of Algorithm \ref{alg:mpc}.}
    \label{tab:mpc}
    \vspace{-3pt}
    \setlength{\arrayrulewidth}{1.5pt}
    \begin{tabular}{c c}
        \hline
        Hyperparameters & Value \\
        \hline
        Horizon ($H$) & 10 \\
        MPC iterations ($N$) & 6 \\
        Num. samples ($M$) & 500 \\
        Num. policy samples ($M_\pi$) & 30 \\
        Num. elite samples ($M_{elite}$) & 60 \\
        Discount factor ($\gamma$) & 0.99 \\
        Constraints weight ($\lambda$) & 1.0 \\
        Momentum factor ($\beta$) & 0.95 \\
        Temperature ($T$) & 0.5 \\
        \hline
    \end{tabular}
    \vspace{-3pt}
\end{table}
}
Solving the optimization problem in \eqref{eq:inf_mpc} is NP-hard. To tackle this, there are various online optimization techniques, like sampling-based approaches, which offer speed advantages over gradient or Hessian-based methods. In this work, we adopt a simple constrained Model Predictive Path Integral (MPPI) \cite{mppi} approach, as detailed in Algorithm \ref{alg:mpc}, with its hyperparameters listed in Table \ref{tab:mpc}. However, our framework is generic and can accommodate various other Model Predictive Control (MPC) algorithms such as those shown in \cite{dmdmpc, rilqr, d3p}, with each having its pros and cons.

As stated earlier, we choose to use the Dreamer of NLM variant in Algorithm \ref{alg:mpc} due to its lower complexity and reduced inference cost compared to PLM and FLM. Recall that, unlike other models, NLM does not rely on any history of observations or latent spaces, which is the reason for its simplicity in implementation. Additionally, planning in latent spaces can compromise interpretability and limit our ability to enforce domain-specific constraints. In the next section, we comprehensively discuss our results with PIP-Loco.

%% file: sections/results.tex
\section{Results}\label{sc:results}

This section details the simulation and hardware experiments with PIP-Loco. We also present ablation studies on the internal model to validate the importance of a $H$ step Dreamer. For training, we utilized an open-source environment for the Unitree Go1 quadruped robot, which is based on Nvidia's Isaac Gym simulator \cite{isaac_gym} \cite{leggedgym}, and implemented all neural networks using the PyTorch framework \cite{pytorch}. The training process was executed on a desktop system with an AMD Ryzen Threadripper Pro 5995WX processor (128 cores) running at 2.7 GHz, 64 GB of RAM, and an NVIDIA RTX 3080 GPU. For real-time planning, we deployed an onboard computer using Jax  \cite{jax}, achieving an inference frequency of 300-500 Hz.

\begin{table}[htp!]
    \scriptsize
    \centering
    \captionsetup{font=small}
    \caption{Performance comparison across different methods and noise levels after 2k iterations for five seeds}

    \begin{subtable}[b]{0.48\textwidth}
        \centering
        \scriptsize
        \setlength{\arrayrulewidth}{1pt}
        \begin{tabular}{c | c c c}
            \hline
            Horizon Length/ Method & 1 & 5 & 10 \\
            \hline
            HIMLoco & 8.59 $\pm$ 0.30 & - & -\\
            PIP-Loco w/ NLM (ours) & \textbf{9.52 $\pm$ 0.38} & {9.69 $\pm$ 0.28} & \textbf{9.58 $\pm$ 0.60}\\
            PIP-Loco w/ PLM (ours) & 9.37 $\pm$ 0.62 & \textbf{9.75 $\pm$ 0.54} & 8.80 $\pm$ 0.29\\
            PIP-Loco w/ FLM (ours) & 8.22 $\pm$ 0.38 & 8.13 $\pm$ 0.83 & 8.84 $\pm$ 0.48\\
            \hline
        \end{tabular}
        \vspace{2mm}        
        \caption{\textit{Final return comparison with different horizons}: The table compares final returns for HIMLoco and PIP-Loco (NLM, PLM, FLM). PIP-Loco consistently outperforms HIMLoco, with PIP-Loco (NLM, H=5) achieving the highest return with the least variance.}
        \label{tab:horizon_result}
    \end{subtable}
    \vspace{2mm}
    \begin{subtable}[b]{0.48\textwidth}
        \scriptsize
        \centering
        \setlength{\arrayrulewidth}{1pt}
        \begin{tabular}{c | c c c}
            \hline
            Noise Level/ Method & Low & Medium & High \\
            \hline
            HIMLoco & 8.96 $\pm$ 0.24 & 8.79 $\pm$ 0.72 & 1.73 $\pm$ 5.88\\
            PIP-Loco w/ NLM (ours) & 9.62 $\pm$ 0.30 & \textbf{8.94 $\pm$ 0.48} & \textbf{8.16 $\pm$ 1.13}\\
            PIP-Loco w/ PLM (ours) & \textbf{9.84 $\pm$ 0.55} & 8.23 $\pm$ 0.33 & 7.99 $\pm$ 0.37\\
            \hline
        \end{tabular}    
        \vspace{2mm}
        \caption{\textit{Final return comparison under different noise levels}: The table shows PIP-Loco's robustness across noise conditions (Low, Medium, High). PIP-Loco (NLM, H=5) excels in high-noise scenarios, outperforming HIMLoco, while PLM (H=5) performs competitively across all noise levels.}
        \label{tab:noise_result}
    \end{subtable}
    \vspace{-10pt}
\end{table}

\begin{figure}[htbp!]
    \small
    \centering
    \captionsetup{font=small}
    \includegraphics[width=0.8\linewidth]{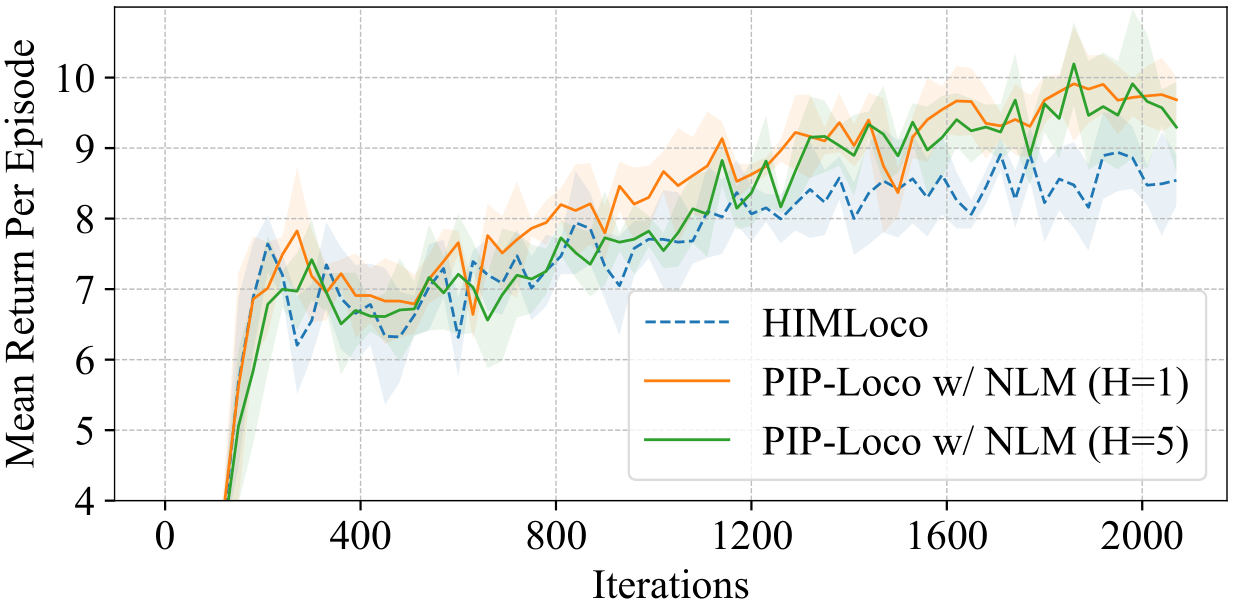}
    \caption{\textit{Training performance comparison}: Mean return per episode over training iterations for PIP-Loco (NLM) at different horizons (H=1, H=5) compared to the baseline HIMLoco \cite{himloco}. PIP-Loco (NLM) shows improved performance across all horizons.}
    \label{fig:reward}
\end{figure}

\begin{figure*}[htp!]
    \small
    \captionsetup{font=small}
    \centering
    \includegraphics[width=0.7\linewidth]{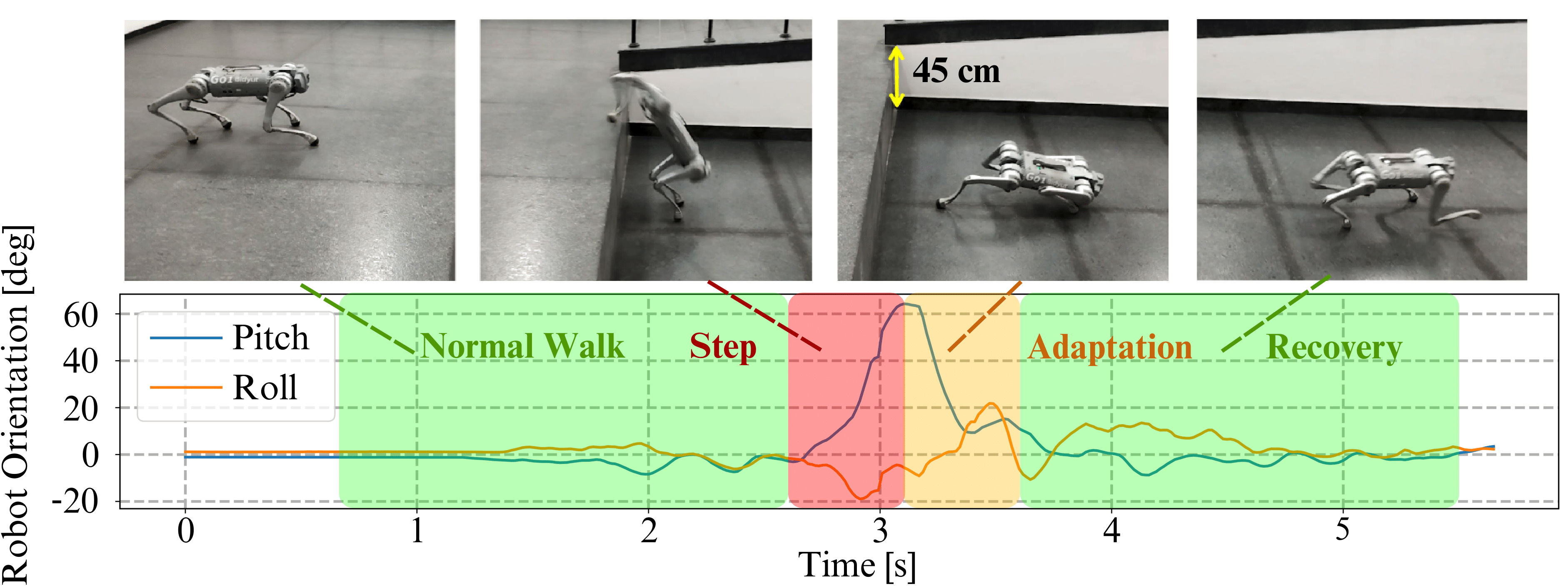}
    \caption{Top Sequence: Robot navigating a 45 cm step-down obstacle, with phases of walking, stepping down, adapting, and recovering. The graph below tracks pitch and roll angles over time, with phases highlighted in green (Normal Walk), red (Step), orange (Adaptation), and green (Recovery).}
    \label{fig:hardware}
    \vspace{-10pt}
\end{figure*}

\subsection{Abalations and Experiments}\label{sc:results-A}

We conducted parallel training of 4096 robots across various terrains, employing domain randomization and a curriculum learning strategy. The PPO algorithm \cite{ppo} used a clipping range of 0.2, a generalized advantage estimation factor of 0.95, and a discount factor of 0.99. All networks were optimized using the Adam optimizer \cite{adam} with a learning rate 0.001.

\begin{figure}[htp!]
    \small
    \captionsetup{font=small}
    \centering
    \begin{subfigure}{0.45\textwidth}
        \centering
        \includegraphics[width=\linewidth,]{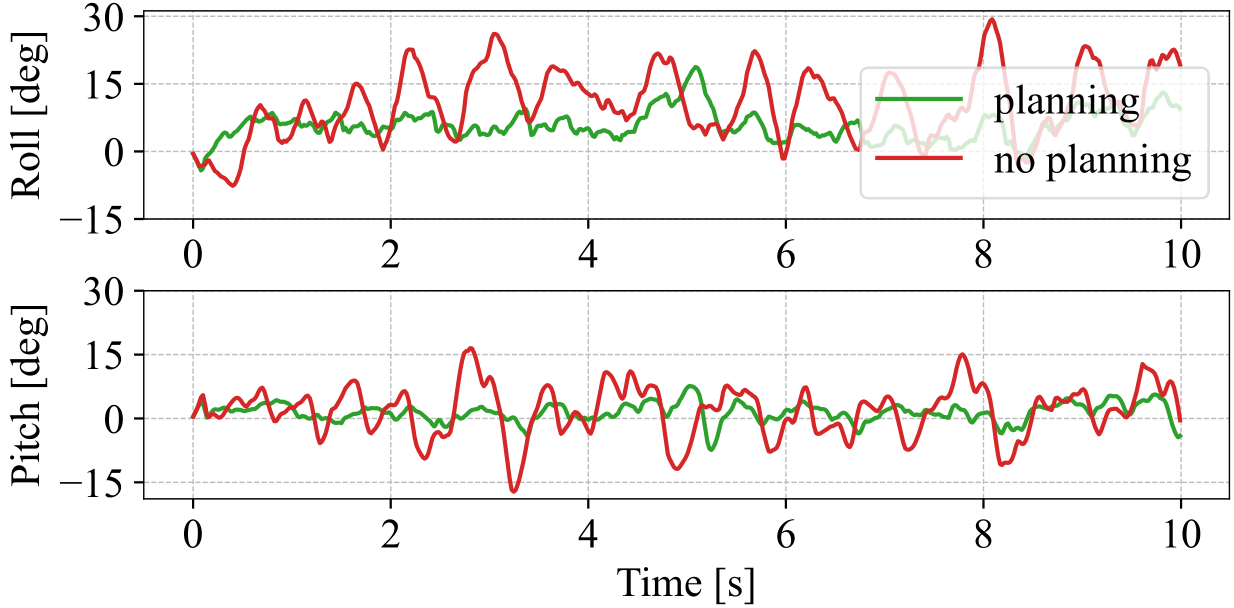}
        \label{fig:plan_vs_pol_rp}
    \end{subfigure}
    \hfill
    \begin{subfigure}{0.45\textwidth}
        \centering
        \includegraphics[width=\linewidth]{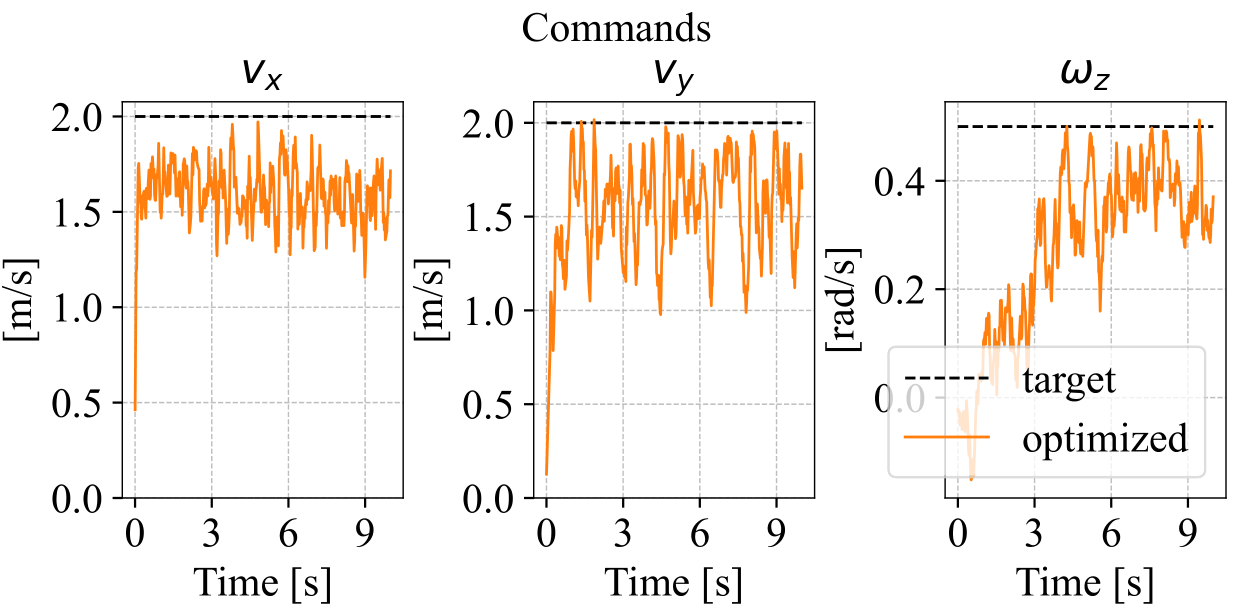}
        \label{fig:plan_vs_pol_cmd}
    \end{subfigure}
    \begin{subfigure}{0.45\textwidth}
        \centering
        \includegraphics[width=\linewidth]{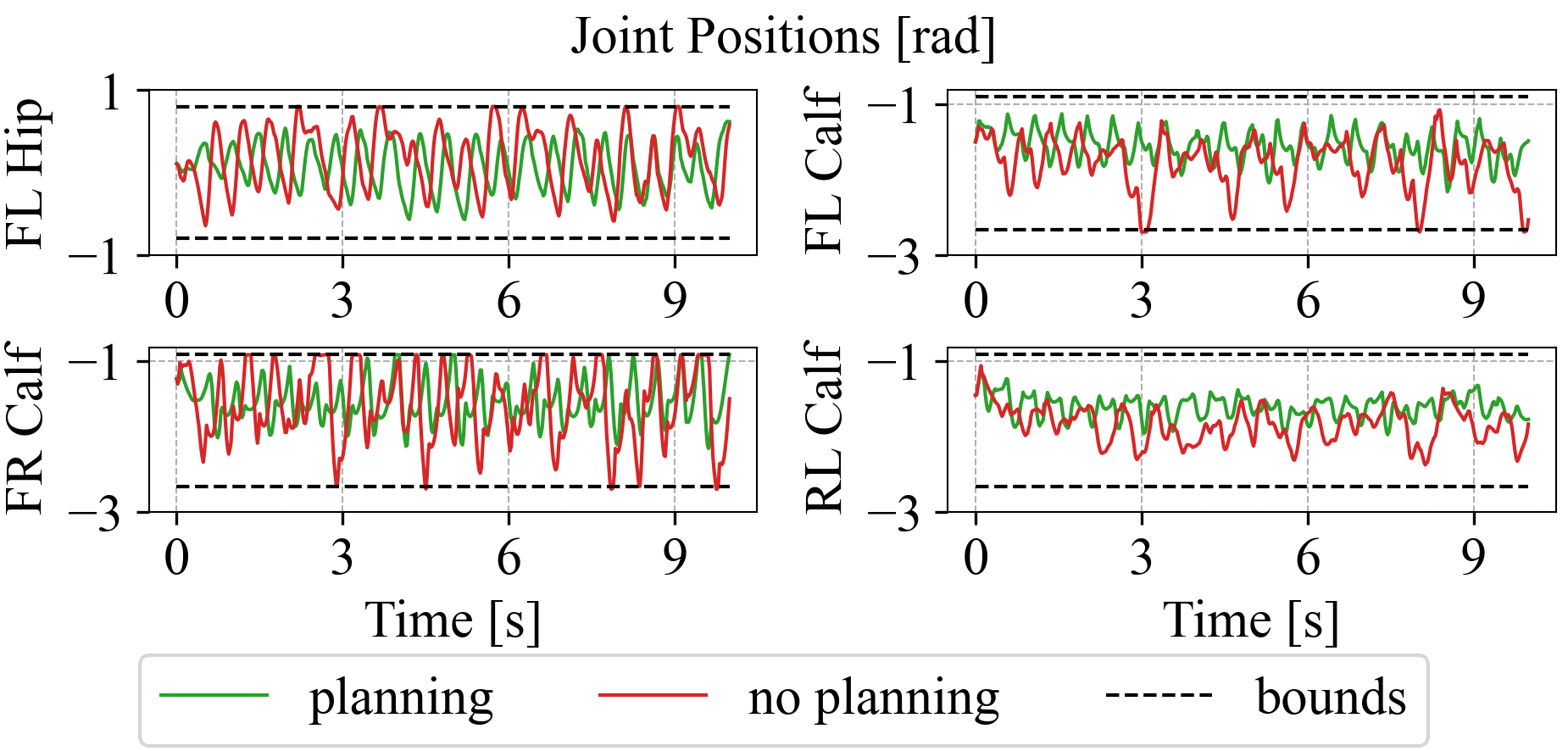}
        \label{fig:plan_vs_pol_jp}
    \end{subfigure}
        
    \caption{\textit{Simulation results for a 10-second run with extreme command velocities:} Top-left compares roll and pitch with and without planning. The top-right shows optimized velocity commands from Algorithm \ref{alg:mpc}. The bottom panel shows joint angle variations, with planning having lower peaks near constraint bounds.}
    \label{fig:plan_vs_policy}
    \vspace{-20pt}
\end{figure}

\subsubsection{Why No MPC During Training?}\label{sc:why_not_offpol}
A natural question that arises is why we do not adopt model-based RL methods for planning, such as those proposed in \cite{loop, mopac, dmdmpc, tdmpc, tdmpc2}, where MPC and RL mutually assist each other during training to enhance performance. While these approaches have demonstrated promise in various contexts and are state-of-the-art in sample efficiency, they pose significant challenges when applied to our setting. In particular, solving the MPC optimization problem in massively parallel RL environments is computationally demanding, leading to considerable bottlenecks in inference speed and memory consumption, thereby significantly increasing the wall clock time. This makes it impractical to adopt these methods directly without substantial modifications. Consequently, in our approach, we defer the planning component to the deployment phase to fully leverage the benefits of planning while avoiding these computational challenges in training.


\subsubsection{What Constitutes a Good Internal Model?}\label{sc:which_internal_model} 
As discussed in \cite{himloco}, an internal model was introduced to simulate a single-step disturbance response in latent space. This work extends that approach by incorporating dreaming of future observations over $H$ steps. Figure \ref{fig:reward} illustrates the reward curves for both the model proposed in \cite{himloco} and our proposed internal models. Additionally, we investigate the influence of the prediction horizon $H$, the model's robustness to noise, and the role of latent states within our framework. The corresponding results are summarized in Tables \ref{tab:horizon_result} and \ref{tab:noise_result}. Our experimental results suggest that the inclusion of latent states does not improve the robustness of the expert actor $\pi_e$ in the presence of training noise. 
We also observed that base velocity estimation is critical to the overall performance for providing accurate dynamics feedback.

\subsubsection{Advantages of Planning} 
We conducted an experiment to assess the robot's ability to track extreme velocity commands, which can result in unstable behaviors. Fig. \ref{fig:plan_vs_policy} presents a comparative analysis between the cloned policy $\pi_\theta$, which operates without planning, and Algorithm \ref{alg:mpc}, which incorporates planning. The experiment aimed to track a target velocity of $\text{v}_x^{cmd} = 2.0 \si{\ \meter / \second}, \text{v}_y^{cmd} = 2.0 \si{\ \meter / \second}, \omega_z^{cmd} = 0.5 \si{\ \radian / \second}$. The planning mechanism in PIP-Loco successfully optimized actions and velocity commands to minimize orientation deviations and maintain base stability while ensuring that joint positions have lower peaks that hit the constraint limits. 
As a future work, we aim to integrate obstacle avoidance and navigation-related constraints in the planner.
The experimental results demonstrate the need for planning in online adaptability.
It is important to note that by integrating planning, we could add/remove constraints and reformulate reward/cost functions in real time without retraining. This flexibility is a key characteristic of MPC approaches, which has been notably absent in RL methods for legged locomotion.

\subsubsection{Hardware Deployment}
We evaluated our planner in various real-world environments, including slopes, gravel surfaces, and stairs. Additional results can be found in the accompanying video submission. Figure \ref{fig:hardware} illustrates an experiment where the robot was commanded to traverse a steep drop of $45 \si{\centi \meter}$. During this process, the robot maintains a stable orientation in the normal walking phase, experiences notable deviations in pitch and roll during the step-down, and subsequently adjusts and recovers to restore stable walking.

%% file: sections/conclusion.tex
\section{Conclusion}

We introduce PIP-Loco, a novel proprioceptive infinite-horizon planning framework for quadrupedal robot locomotion. Our approach overcomes the limitations of RL-based controllers in terms of interpretability and safety while addressing the task complexity and finite-horizon issue typical of classical MPC methods. By integrating our proposed internal model, PIP-Loco enhances the exploration of effective locomotion behaviors during training and solves an MPC problem at deployment.
Extensive ablation studies and multi-terrain tests in both simulation and hardware confirm its robustness and real-world potential.

However, a key limitation is the reliance on the assumption that the expert RL policy and the sampling-based planner behave similarly. Significant deviations during deployment may cause the internal model trained on expert actor behavior to generate unreliable predictions. Future work will explore the integration of learning-based safety filters to align the planner with policy behavior better, as well as incorporate advancements in perceptive locomotion \cite{parkoureth} and mitigate dynamics prediction errors.
